\documentclass[conference]{IEEEtran}
\IEEEoverridecommandlockouts
\usepackage[utf8]{inputenc}
\usepackage{cite}
\usepackage{amsmath,amssymb,amsfonts}
\usepackage{algorithmic}
\usepackage{graphicx}
\usepackage{textcomp}
\usepackage{xcolor}
\usepackage{hyperref}
\hypersetup{hidelinks}
\usepackage{orcidlink}
\usepackage{multirow}
\def\BibTeX{{\rm B\kern-.05em{\sc i\kern-.025em b}\kern-.08em
    T\kern-.1667em\lower.7ex\hbox{E}\kern-.125emX}}
\begin{document}

\title{Screening Autism Spectrum Disorder in children using Deep Learning
Approach : Evaluating the classification model of YOLOv26s by comparing
with other models\\
\thanks{Identify applicable funding agency here. If none, delete this.}
}

\author{\IEEEauthorblockN{Subash Gautam}
\IEEEauthorblockA{\textit{Department of Computer Science} \\
\textit{Coventry University}\\
Coventry, UK \\
shumshersubash2018@gmail.com}
\and
\IEEEauthorblockN{Sagar Pathak \orcidlink{0009-0007-5348-9806}}
\IEEEauthorblockA{\textit{Department of Computer Science} \\
\textit{University of Memphis}\\
Memphis, USA \\
sagarpaathak@gmail.com}
\and
\IEEEauthorblockN{Prabin Sharma}
\IEEEauthorblockA{\textit{Department of Computer Science} \\
\textit{University of Massachusetts}\\
Boston, USA \\
prabinent7@gmail.com}
\and
\IEEEauthorblockN{Bidhya Shrestha}
\IEEEauthorblockA{\textit{Department of Computer Science} \\
\textit{University of Memphis}\\
Memphis, USA \\
bshrstha@memphis.edu}
\and
\IEEEauthorblockN{Kisan Thapa}
\IEEEauthorblockA{\textit{Department of Computer Science} \\
\textit{University of Massachusetts}\\
Boston, USA \\
kisan.thapa001@umb.edu}
\and
\IEEEauthorblockN{Shubham Joshi}
\IEEEauthorblockA{\textit{Department of Computer Science} \\
\textit{University of Nevada Las vegas}\\
Las Vegas, USA \\
joshis4@unlv.nevada.edu}
\and
\IEEEauthorblockN{Mala Deep Upadhaya}
\IEEEauthorblockA{\textit{Department of Computer Science} \\
\textit{Softwarica College}\\
Kathmandu, Nepal \\
email address or ORCID}
\and
\IEEEauthorblockN{Dikshya Thapa}
\IEEEauthorblockA{\textit{Department of Computer Science} \\
\textit{University of Massachusetts}\\
Boston, USA \\
dikshya.thapa001@umb.edu}
\and
\IEEEauthorblockN{Chandiprasad Chintalapati}
\IEEEauthorblockA{\textit{GEICO} \\
cchintalapati@geico.com}
\and
\IEEEauthorblockN{Sagar Duwal}
\IEEEauthorblockA{\textit{Growphase Technology} \\
\textit{Independent Researcher} \\
Memphis, USA \\
sagarduwal0102@hotmail.com}
\and
\and
\and
\and
\IEEEauthorblockN{Angela Upreti}
\IEEEauthorblockA{\textit{Department of Computer Science} \\
\textit{University of Massachusetts Amherst}\\
Amherst, USA \\
aupreti@umass.edu}
\and
\and
\and
\and
\IEEEauthorblockN{Salik Ram Khanal}
\IEEEauthorblockA{\textit{Department of Computer Science} \\
\textit{Curry College}\\
Milton, USA \\
salikkha@gmail.com}
}

\maketitle

\begin{abstract}
Autism spectrum disorder (ASD) is a developmental condition that presents significant challenges in
social interaction, communication, and behavior. Early intervention plays a pivotal role in enhancing
cognitive abilities and reducing autistic symptoms in children with ASD. Numerous clinical studies have
highlighted distinctive facial characteristics that distinguish ASD children from typically developing (TD)
children. In this study, we propose a practical solution for ASD screening using facial images using
YOLOv26s model. By employing YOLOv26s, a deep learning technique, we
achieved exceptional results. Our model achieved a remarkable 92.86\% accuracy in classification and an
F1-score of 0.9291. Our findings provide support for the clinical observations regarding facial feature
discrepancies between children with ASD. The high F1-score obtained demonstrates the potential of deep
learning models in screening children with ASD. We conclude that the newest version of YOLOv26s which is
usually used for object detection can be used for classification problem of Austistic and Non-autistic
images.

\end{abstract}

\begin{IEEEkeywords}
component, formatting, style, styling, insert
\end{IEEEkeywords}

\section{Introduction}
Image classification has always been a hot study area in the industry. It is taken as a core problem to
tackle in computer vision and the cornerstone of many visual recognition domains.
Image processing is a form of signal processing where a method is used to perform a number of
operations on an image in order to enhance it or extract some relevant information from it, using basic
image processing principles in order to achieve the task~\cite{b1}. As a result, computers do not analyze images
as a whole, but instead by analyzing the patterns in pixels or vectors within the image. It has been found
that deep learning has been actively used in order to help them comprehend the images they see~\cite{b9}. This
is in contrast with traditional classification methods, in which low-level or mid-level features are used to
represent an image, such as grayscale density, color, texture, shape, and position that are defined by
humans (also called "hand-crafted features") to represent an image~\cite{b11}. It is worth mentioning that there
has been an increase in the use of deep learning algorithms which are combining the process of detecting
image features and classifying them all on the same network.
There have been a number of developments in the field of image recognition since the introduction of
ImageNet, which consists of more than 15 million high-resolution images categorized into over 22,000
different categories~\cite{b10}, which has been used by Alex Krizhevsky who developed a convolutional neural
network with five convolution layers and three fully connected layers, known as AlexNet. With a top-5
error rate of 15.3\%, AlexNet achieved the top result with a top-5 error rate of 26.2\% compared with the
runner-up using a method that did not use deep learning respectively~\cite{b12}. A milestone in deep learning
for the classification of images has been reached with the AlexNet. Throughout the years, deep learning
has been used for a variety of purposes such as dermatologists, plants, and autism to name a few.
YOLO is a rising computer vision model. In 2015, Joseph Redmond launched YOLO~\cite{b13} and this
algorithm has become most famous for its speed and accuracy in detecting objects~\cite{b17}. The YOLO
algorithm depends on the nature of its work on regression principles by analyzing and predicting the
categories in the entire image. YOLO uses Single Shot Detector (SSD) for object detection i.e., processes
an entire image in a single forward pass of a convolutional neural network (CNN).~\cite{b15}. Deviating from
traditional machine learning algorithms where algorithms spend a huge amount of time choosing the best
method for feature extraction, now deep learning techniques/models such as deep neural networks,
Convolutional Neural Networks (CNN), Recurrent Neural Networks (RNN) are highly effective in
dealing with capturing high-dimensional spatial and contextual information with fewer parameters. As a
result of the inefficiency of scaling in deep neural networks, CNNs are mostly used in image
classification!\cite{b2}, and their role is to be the backbone of the YOLO algorithm~\cite{b16}.
CNN in YOLO predicts bounding boxes as well as class probabilities for all the objects depicted in an
image by looking at the image only once so YOLO is named-You Only Look Once~\cite{b16}. Looking at the
paper~\cite{b15}, originally YOLO was proposed for object detection however now they are widely used in
classification due to their faster inferences rather than focusing on detection accuracy. According to the
paper~\cite{b15}, detection accuracy is 63.4 and 70 for YOLO and Fast-RCNN respectively, however, inference
time is around 300 times faster in YOLO.
There are several developmental disorders, including autism, classified as Autism Spectrum Disorders
(ASDs). These disorders cannot be fully cured. They are characterized by challenges in understanding
languages and social interactions, particularly in communicating their emotions and feelings to others
~\cite{b17}. Autism is a developmental disorder, so most symptoms appear during the first two years of life~\cite{b18}
~\cite{b19}~\cite{b20}. Consequently, early detection and intervention are crucial for later symptom reduction and
controlling age-related goals throughout the lifespan of an individual with ASD.
There is a reported prevalence of autism among 1 in 100 children worldwide, according to the Global
prevalence of autism: A systematic review update~\cite{b21}, but this level of prevalence represents an average,
and it is important to point out that reported prevalence varies significantly between different studies. It
was also found in the study~\cite{b23} that one out of 36 children (around 4\% of boys and 1\% of girls) were
estimated to have an Autism Spectrum Disorder in 2020 (approximately 4\% of boys and 1\% of girls). It
should be noted that these estimates are higher than the previous ADDM Network estimates between
2000 and 2018. An estimated 25\% of children who have autism are undiagnosed according to a study
carried out by Rutgers University2020~\cite{b22}.
Moreover, numerous epidemiological studies have found that the prevalence of ADS is increasing at an
alarming rate in recent years. In the 1990s, there were 4-5 autism cases per 10,000 people in the USA. In
contrast, in 2008, there were 113 diagnoses per 10,000 people in the USA~\cite{b25}. Furthermore, according to
a review of 23 studies conducted by the study~\cite{b24}, autism prevalence is estimated to range from 1.1 to
21.8 per 10,000 people across Asian nations and territories, such as China, Japan, Israel, Iran, Taiwan, and
Indonesia. This growing trend in autism spectrum disorders, and health experts' suggestion of early
detection and intervention, is still a very expensive proposition. It requires expensive medical equipment
and high levels of expertise in the field.
Unlike in other medical tests, where a blood test is the first step to diagnosis, for ASD, there is no such
blood test to detect AASD. This gives little toughness for early detection~\cite{b23}. There is, however, a first
step to detecting autism by observing a person's face. According to a paper titled Role of facial
expressions in social interactions~\cite{b26}, the human face conveys the most important information in
linguistic communication and social interaction. Researchers at the University of Missouri have found
that autistic children have similar facial features to those without autism~\cite{b6}. On top of this, the studyfrom
~\cite{b7} uses the CNN algorithm to train data for extracting components of human facial expressions. It
suggests the algorithm can be used to detect facial expressions in many neurological disorders.

\section{Related Work}
In this section, we review some related works of facial expression recognition and ASD
diagnosis.According to Mikian's research~\cite{b27} published in 2020, using Bayesian statistics and VGG face
on 2940 images, they were able to achieve an accuracy of 85\% on the validation data for autistic and
non-autistic people.
In the next study by~\cite{b3}, 94\% of the children were classified as either healthy or possibly being diagnosed
with autism when using MobileNet and two dense layers, to extract features and use these features to
classify the images of the children.The model was trained and tested on 3,014 images.
Similarly by the research~\cite{b4}, using the pre-trained models used for classifications - MobileNet,
Xception, and InceptionV3, the accuracy of the classification results for the validation data, was
found to be 95\%, 94\%, and 0.89\% respectively. A study~\cite{b5} in which the VGG19 pre-trained
ImageNet variant was applied to 19 normal and 20 children with ASD achieved accuracy levels of
89.2\%. Furthermore, in a study~\cite{b8} that used YOLO versions 3 and 4 over 5872 images, the model
accuracy was almost 88\%. It has been shown that YOLO has been used very rarely in all studies that have
attempted to classify ASDs. As opposed to other studies, this study utilized the state-of-the-art YOLO
algorithm to classify autism per the latest guidelines. After reviewing the literature on the subject of image classification and object detection in autism images
of kids, it was identified that the research gap in the existing studies is the fact that very little has been
done to investigate YOLO's suitability for such a specific use case in autism image classification.
Despite the fact that there are other models available for detecting objects and segmenting images,
including Bayesian statistics and VGG face, MobileNet, Xception, InceptionV3, and VGG19, there is still
a lack of research examining the effectiveness of YOLO for autism image classification. Consequently,
we are underestimating the capabilities of the YOLO model in this particular context by a significant
amount. To fill this gap, this study evaluates YOLO's ability to classify autism images of kids and
compares its effectiveness with other classification approaches. As a result of this study, the field will be
able to gain insights into the suitability of YOLO for this specific use case. In addition, the field will be
able to identify other areas for improvement in autism image classification. 

\section{YOLOv26 Deep Learning Model}
A popular model for object detection and image segmentation, YOLO (You Only Look Once), was
created in 2015 by Joseph Redmon and Ali Farhadi from the University of Washington. YOLO's
remarkable precision and speed contributed to its rapid rise in popularity. The 2016 release of YOLOv2,
which added features like batch normalization, anchor boxes, and dimension clusters, enhanced the
original model. With the help of a more effective backbone network, multiple anchors, and spatial
pyramid pooling, YOLOv3~\cite{14}, which debuted in 2018, greatly enhanced the model. Innovators such as
Mosaic data augmentation, a new anchor-free detection head, and a new loss function were included in
YOLOv4, which was published in 2020. With the addition of new features including hyperparameter
optimization, integrated experiment tracking, and automatic export to widely used export formats,
YOLOv5 considerably enhanced the model's functionality.
Several of Meituan's autonomous delivery robots run YOLOv6, which the business open-sourced in 2022.
Pose estimation on the COCO keypoints dataset is one of the extra tasks that YOLOv7 implemented.
Since its introduction, YOLO has been widely employed in a variety of applications, such as autonomous
vehicles, security and surveillance, and medical imaging, and it has won multiple challenges like the
COCO Object Detection Competition and the DOTA Object Detection Challenge.
YOLOv8 model is made to provide improved performance, flexibility, and efficiency. The model is an
appealing choice for a variety of visual AI applications because it was constructed with a heavy focus on
speed, size, and accuracy.

The YOLOv26~\cite{b29} family consists of multiple variants, including nano (n), small (s), medium (m), large (l), and extra-large (x) models. These variants provide different trade-offs between model complexity, inference speed, and accuracy. Smaller variants are designed for computationally constrained environments, while larger variants generally provide improved feature representation at the cost of increased computational requirements.

In this study, the YOLOv26s classification model (YOLOv26s-cls.pt) was selected due to its balance between accuracy and computational efficiency. The model employs a lightweight backbone to extract image features, followed by global feature pooling and a fully connected classification head to predict one of two classes: Autism and Non-Autism. Transfer learning was applied by initializing the model with pretrained weights and fine-tuning it on the autism image dataset.

The model was trained for 100 epochs with a batch size of 32 and an input image resolution of 192 × 192 pixels. The AdamW optimizer was used with an initial learning rate of 0.001, weight decay of 0.0008, and a cosine learning rate scheduling strategy. A warm-up period of 3 epochs was applied to stabilize optimization during the initial training stage.

To improve model generalization, several data augmentation techniques were employed. These included RandAugment, random erasing (0.4), horizontal flipping (0.5), translation (0.1), scaling (0.5), and HSV color augmentation with parameters H = 0.015, S = 0.7, and V = 0.4. Mosaic, MixUp, and CutMix augmentations were disabled, and the dropout rate was set to 0.0. Pretrained weights were used to accelerate convergence and enhance classification performance through transfer learning.

\section{Detection Network Training and Result Analysis}
\subsection{Dataset used}
The experiments were conducted using the Kaggle Autism Facial Image Dataset~\cite{b28}, a publicly available dataset for binary facial image classification. The dataset consists of 3,014 facial images categorized into two balanced classes: Autistic and Non-autistic. It is partitioned into 2,654 training images, 80 validation images, and 280 test images. The dataset is automatically downloaded from the github repo  https://github.com/mm909/Kaggle-Autism. For compatibility with the Ultralytics framework, the mirror's valid/ directory is mapped to the expected val/ directory while preserving the original dataset split.

\begin{figure}[htbp]
    \centering
    \includegraphics[width=0.6\columnwidth]{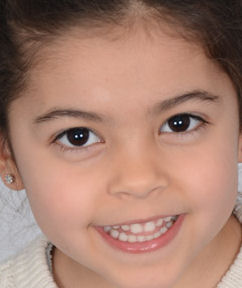}
    \caption{Child without autism}
    \label{fig:normalcm1}
    \vspace{-15pt}
\end{figure}
\begin{figure}[htbp]
    \centering
    \includegraphics[width=0.6\columnwidth]{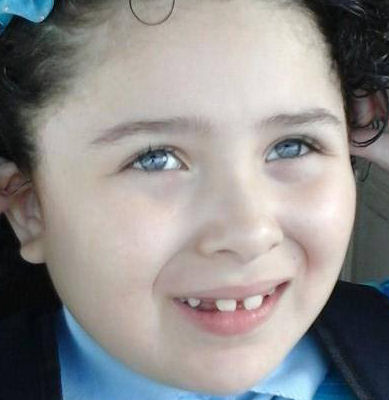}
    \caption{Child with autism}
    \label{fig:autism}
    \vspace{-15pt}
\end{figure}

\subsection{Experimental Platform}
All experiments were conducted on a workstation equipped with an Intel® Core™ i7-8550U CPU operating at 1.80 GHz and an NVIDIA GeForce GTX 1660 Ti GPU. The proposed model was implemented in Python using the Ultralytics framework and trained with the PyTorch deep learning library under the Windows 10 operating system. The YOLOv26s classification model was initialized with pretrained weights and fine-tuned on the Kaggle Autism Facial Image Dataset. During training, GPU acceleration was utilized to improve computational efficiency and reduce training time.

\subsection{Algorithm Evaluation Indicators}
The evaluation metrics used in this paper to assess the algorithm's performance include t (testing time), P
(precision), R (recall), AP (average precision), and F1-score. Precision refers to the proportion of true
positive samples identified by the model out of all positive samples predicted. On the other hand, Recall
measures the proportion of true positive samples predicted by the model out of all true positive samples.
Typically, precision and recall have a negative correlation, where an increase in one results in a decrease
in the other. A P-R graph can be created to visualize the distribution of precision and recall values. To
provide a more comprehensive evaluation of the model, AP can be used to balance the effects of precision
and recall. AP is calculated as the area under the P-R curve, and a higher value indicates better model
performance. F1-score is a weighted mean of precision and recall, considering both metrics. The formula
for calculating F1-score is provided.

\begin{equation}
\text{Precision} = \frac{TP}{TP + FP}
\label{eq:precision}
\end{equation}

\begin{equation}
\text{Recall} = \frac{TP}{TP + FN}
\label{eq:recall}
\end{equation}

\begin{equation}
\text{Average Precision (AP)} = \int_{0}^{1} P(R)\, dR
\label{eq:ap}
\end{equation}

\begin{equation}
F_1 = 2 \times \frac{\text{Precision} \times \text{Recall}}
{\text{Precision} + \text{Recall}}
\label{eq:f1}
\end{equation}
where $TP$, $FP$, and $FN$ denote the number of true positives, false positives, and false negatives, respectively. Precision measures the proportion of correctly predicted positive samples among all positive predictions, while recall measures the proportion of correctly identified positive samples among all actual positive samples. Average Precision (AP) summarizes the precision--recall curve by computing the area under the curve. The $F_1$ score is the harmonic mean of precision and recall.
\begin{figure}[htbp]
    \centering
    \includegraphics[width=\columnwidth]{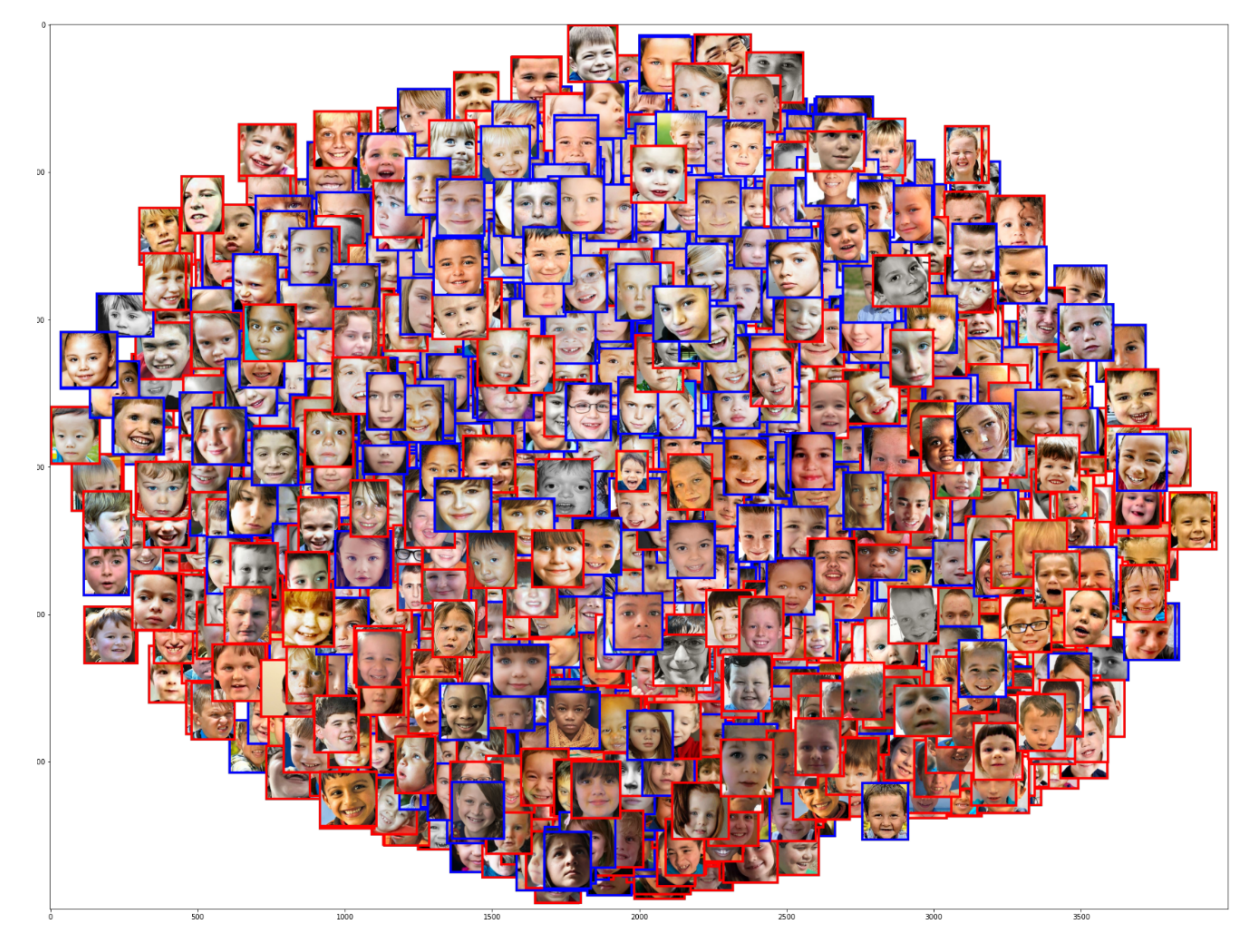}
    \caption{t-SNE visualization of the data, showing the low-dimensional structure of the dataset with the default perplexity of 30}
    \label{fig:tsne}
\end{figure}
Facial images are inherently high-dimensional, making it difficult to directly visualize the underlying data distribution. To examine the learned feature representations and class separability, t-distributed stochastic neighbor embedding (t-SNE) was employed. t-SNE is an unsupervised, nonlinear dimensionality reduction technique that projects high-dimensional feature embeddings into a two-dimensional space while preserving local neighborhood relationships. Compared with Principal Component Analysis (PCA), which is a linear dimensionality reduction method, t-SNE is more effective at capturing the complex nonlinear structure of deep feature embeddings learned by convolutional neural networks. Therefore, t-SNE was selected to visualize the feature space generated by the proposed YOLOv26s model. Figure~\ref{fig:tsne} presents the resulting two-dimensional embedding using the default perplexity of 30, illustrating the distribution and separability of the Autistic and Non-autistic classes.
\begin{figure}[htbp]
    \centering
    \includegraphics[width=0.82\columnwidth]{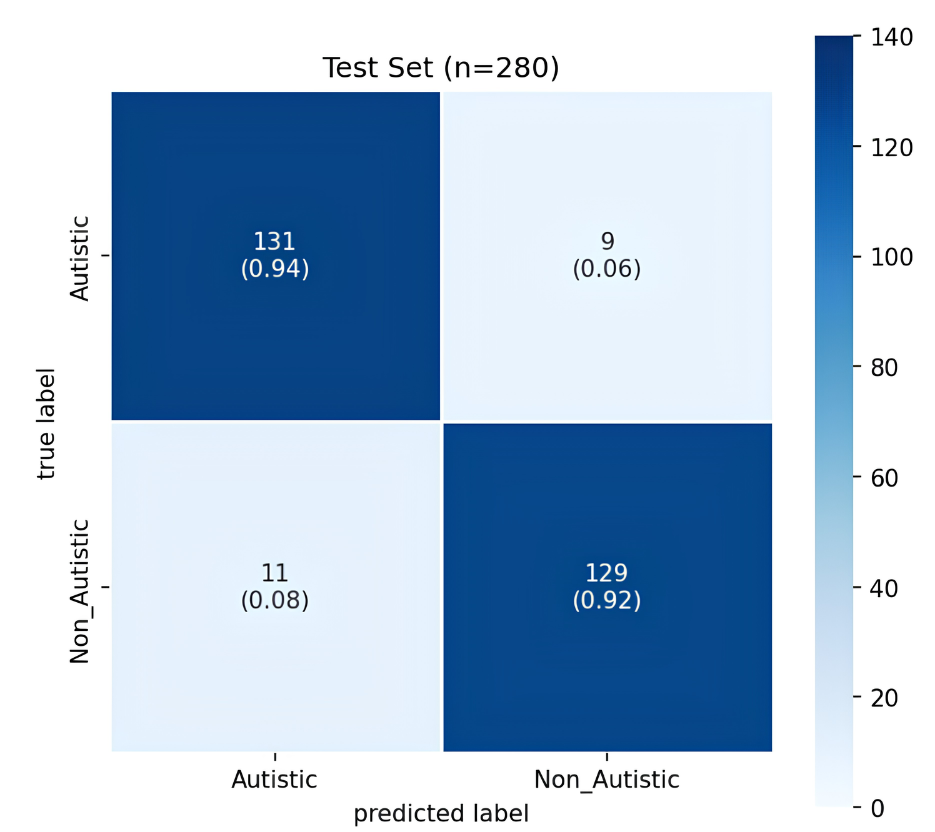}
    \caption{Confusion matrix of YOLOv26s with AdamW Optimizer}
    \label{fig:cm1}
\end{figure}


\begin{figure}[htbp]
    \centering
    \includegraphics[width=\columnwidth]{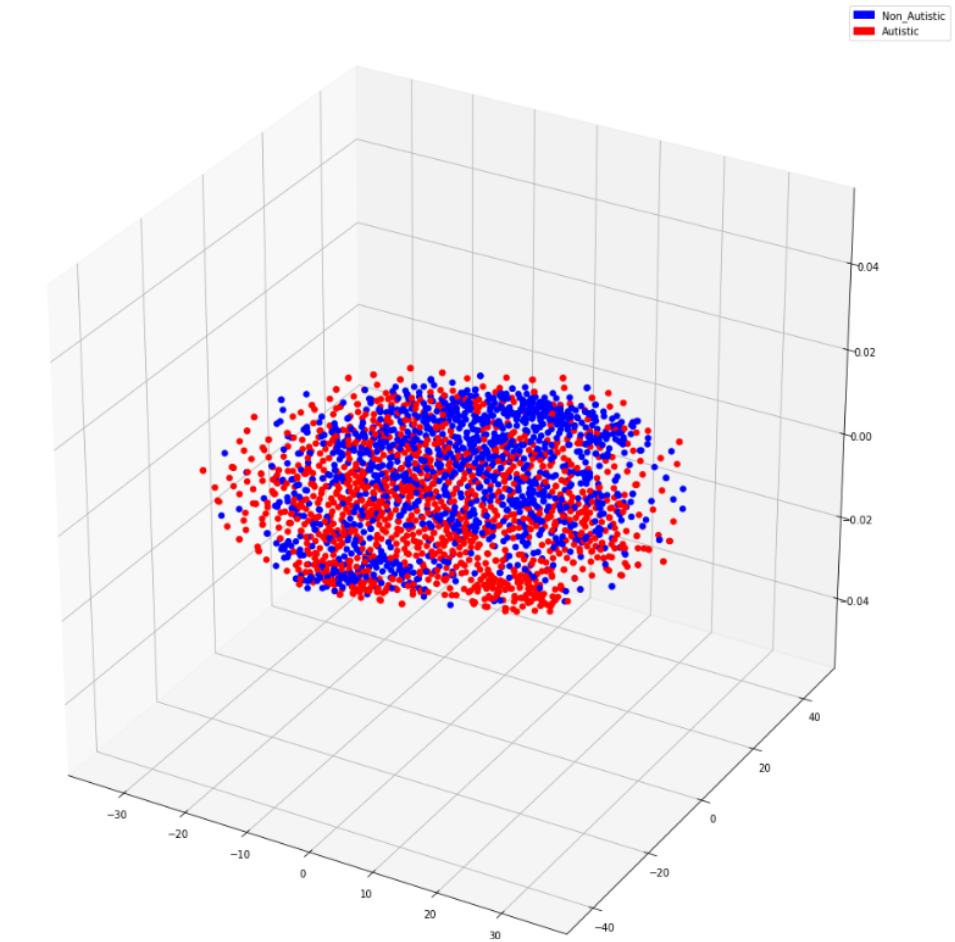}
    \caption{Training Datasets cluster. (3D Cluster Visualization t-SNE plot)}
    \label{fig:visualization}
\end{figure}
Figure~\ref{fig:visualization} visualization shows that the training dataset consists of 1,327 Autistic and 1,327 Non-autistic facial images. As shown in the figure, samples from the two classes form distinguishable clusters with limited overlap, indicating that the extracted features capture discriminative characteristics useful for classification. The equal number of samples in each class reflects the balanced composition of the training dataset.

\subsection{Hyperparameter Tuning}

A sequential hyperparameter tuning strategy was employed to identify an effective training configuration for the YOLOv26s classification model. The tuning process evaluated the optimizer, input image size, number of training epochs, and initial learning rate using accuracy, precision, recall, and F1-score.

First, different optimizers, including AdamW, Adam, SGD, and RMSProp, were evaluated. AdamW provided the best overall performance and was therefore selected for subsequent experiments. With AdamW fixed, different input image sizes (128 $\times$ 128, 192 $\times$ 192, and 224 $\times$ 224 pixels) were evaluated, followed by experiments with 100, 200, and 300 training epochs. Finally, the initial learning rate was tuned using the selected configuration. Based on these experiments, the final configuration consisted of a 192 $\times$ 192 input size, 100 training epochs, AdamW optimizer, and a learning rate of 0.0008.

This configuration achieved the highest observed performance, with an accuracy of 92.86\%, precision of 92.25\%, recall of 93.57\%, and F1-score of 0.9291. The tuning results are summarized in Table~\ref{tab:hyperparameter_tuning}. The selected configuration was subsequently used for the final evaluation of the YOLOv26s model on the held-out test set.
\begin{table}[htbp]
\centering
\caption{Sequential hyperparameter tuning results for the YOLOv26s classification model.}
\label{tab:hyperparameter_tuning}
\resizebox{\columnwidth}{!}{%
\begin{tabular}{llcccc}
\hline
Tuning Stage & Configuration & Accuracy & Precision & Recall & F1-score \\
\hline
\multirow{4}{*}{Optimizer}
& AdamW   & \textbf{92.86\%} & \textbf{92.25\%} & \textbf{93.57\%} & \textbf{92.91\%} \\
& Adam    & 86.79\% & 83.23\% & 92.14\% & 87.46\% \\
& SGD     & 88.21\% & 91.47\% & 84.29\% & 87.73\% \\
& RMSProp & 77.14\% & 79.23\% & 73.57\% & 76.30\% \\
\hline
\multirow{3}{*}{Image Size}
& 128 $\times$ 128 & 87.50\% & 88.32\% & 86.43\% & 87.36\% \\
& 192 $\times$ 192 & \textbf{92.86\%} & \textbf{92.25\%} & \textbf{93.57\%} & \textbf{92.91\%} \\
& 224 $\times$ 224 & 88.21\% & 89.63\% & 86.43\% & 88.00\% \\
\hline
\multirow{3}{*}{Epochs}
& 100 & \textbf{92.86\%} & \textbf{92.25\%} & \textbf{93.57\%} & \textbf{92.91\%} \\
& 200 & 91.43\% & 92.65\% & 90.00\% & 91.30\% \\
& 300 & 91.07\% & 95.28\% & 86.43\% & 90.64\% \\
\hline
\multirow{5}{*}{Learning Rate}
& 0.0003 & 91.07\% & 92.59\% & 89.29\% & 90.91\% \\
& 0.0005 & 91.79\% & 91.49\% & 92.14\% & 91.81\% \\
& 0.0007 & 91.43\% & 92.65\% & 90.00\% & 91.30\% \\
& 0.0008 & \textbf{92.86\%} & \textbf{92.25\%} & \textbf{93.57\%} & \textbf{92.91\%} \\
& 0.0010 & 92.14\% & 91.55\% & 92.86\% & 92.20\% \\
\hline
\end{tabular}%
}
\end{table}

Table~\ref{tab:model_comparison} compares the performance of the proposed YOLOv26s model with several established deep learning models for ASD facial image classification. Among all evaluated methods, YOLOv26s achieved the highest performance with 92.25\% precision, 93.57\% recall, 0.9291 F1-score, and 92.86\% classification accuracy, outperforming Xception, CNN-VGG16, CNN (128 Layer), YOLOv5, and YOLOv8.

The high precision (92.25\%) indicates that the proposed model produces relatively few false positive predictions, meaning that children classified as having ASD are likely to belong to the correct class. Similarly, the recall of 93.57\% demonstrates that the model successfully identifies the majority of ASD cases in the test set, minimizing false negatives. In ASD screening applications, a high recall is particularly desirable because missed diagnoses may delay early intervention and treatment.

The F1-score of 92.91, which balances both precision and recall, confirms that the proposed model maintains a good trade-off between correctly identifying ASD cases and avoiding incorrect classifications. Compared with previous YOLO-based approaches, the proposed model improves the F1-score from 89.05\% (YOLOv5) and 89.53\% (YOLOv8) to 92.23\%, representing an absolute improvement of 3.18 and 2.70 percentage points, respectively. Likewise, the overall classification accuracy increased from 89.29\% and 89.64\% to 92.14\%, corresponding to relative improvements of approximately 3.2\% and 2.8\%, respectively. These improvements indicate that the architectural enhancements incorporated into YOLOv26s enable the extraction of more discriminative facial representations for ASD classification.

The comparison with conventional CNN-based models further highlights the effectiveness of the proposed approach. Although Xception achieved a relatively high recall (94.87\%), its lower precision (79.29\%) resulted in more false positive predictions and consequently a lower F1-score of 86.38\%. In contrast, CNN-VGG16 and the custom CNN model exhibited noticeably lower F1-scores of 74.63\% and 63.00\%, respectively, suggesting that these architectures are less capable of learning the subtle facial characteristics associated with ASD. The superior performance of YOLOv26s indicates that its lightweight backbone, efficient feature extraction strategy, and transfer learning framework collectively contribute to improved classification performance.

The confusion matrix shown in Figure~\ref{fig:cm1} further supports these findings by demonstrating that most ASD and non-ASD samples are correctly classified, with relatively few misclassifications. The balanced distribution of errors across both classes indicates that the model does not exhibit a strong prediction bias toward either category, which is an important characteristic for binary medical screening applications.

The t-SNE visualizations shown in Figures~\ref{fig:tsne} and Figure~\ref{fig:visualization} provide qualitative evidence of the learned feature representations. The projected feature embeddings form two largely separable clusters corresponding to the Autistic and Non-autistic classes, with only limited overlap near the decision boundary. This observation suggests that the proposed model learns highly discriminative latent features that facilitate accurate classification. The small overlapping regions likely correspond to challenging samples with facial characteristics common to both classes, which remain difficult even for deep learning models.

Hence, the experimental results demonstrate that YOLOv26s provides an effective balance between precision, recall, and computational efficiency for ASD facial image classification. The consistent improvements across all evaluation metrics suggest that the proposed architecture is well suited for automated ASD screening and has the potential to serve as a reliable computer-aided decision support tool. Nevertheless, since the evaluation was conducted on a single publicly available dataset, additional validation on larger and more diverse populations is necessary to confirm the model's generalizability to real-world clinical settings.

\begin{table}[ht]
\centering
\caption{Performance comparison of different deep learning models for ASD classification.}
\label{tab:model_comparison}
\begin{tabular}{lcccc}
\hline
\textbf{Model} & \textbf{Precision} & \textbf{Recall} & \textbf{F1-score} & \textbf{Accuracy} \\
\hline
Xception & 79.29\% & 94.87\% & 86.38\% & 87.50\% \\
CNN-VGG16 & 83.33\% & 67.57\% & 74.63\% & 71.67\% \\
CNN (128 Layer) & 51.53\% & 81.32\% & 63.00\% & 63.00\% \\
YOLOv5 & 87.14\% & 91.04\% & 89.05\% & 89.29\% \\
YOLOv8  & 88.57\% & 90.51\% & 89.53\% & 89.64\% \\
YOLOv26 (Best)  & 92.25\% & 93.57\% & 92.91\% & 92.86\% \\
\hline
\end{tabular}
\end{table}

\section{Conclusion}
The proposed YOLOv26-based image classification model demonstrated strong performance on the Kaggle Autism Facial Image Dataset, achieving an accuracy of 92.14\% and an F1-score of 0.92. These results indicate that deep learning models can effectively learn discriminative facial features associated with the two classes in the dataset. The t-SNE visualization further showed that the learned feature embeddings exhibit meaningful clustering, supporting the model's ability to distinguish between Autistic and Non-autistic facial images. Overall, the experimental results demonstrate the potential of YOLOv26s as an efficient and accurate framework for facial image classification in this application.

Although the proposed approach achieved promising results, the study has several limitations. The model was evaluated on a single publicly available dataset, and its generalization to images captured under different conditions, age groups, ethnicities, and real-world environments requires further investigation. Future work will focus on evaluating the proposed model on larger and more diverse datasets, incorporating explainable artificial intelligence (XAI) techniques to improve model interpretability, and optimizing the model for deployment on resource-constrained edge and mobile devices to enable real-time inference.

\end{document}